\documentclass[final]{cvpr}

\usepackage{times}
\usepackage{epsfig}
\usepackage{graphicx}
\usepackage{amsmath}
\usepackage{amssymb}
\usepackage{makecell}
\usepackage{multirow} 
\newcommand{\ignore}[1]{}
\newcommand{\tabincell}[2]{\begin{tabular}{@{}#1@{}}#2\end{tabular}}  %表格自动换行

\usepackage{appendix}

\usepackage[pagebackref=true,breaklinks=true,colorlinks,bookmarks=false]{hyperref}

\def\cvprPaperID{3472} % *** Enter the CVPR Paper ID here
\def\confYear{CVPR 2021}

\begin{document}

%%%%%%%%% TITLE
\title{Alpha-Refine: Boosting Tracking Performance by \\ Precise Bounding Box Estimation}

\author{
Bin Yan$^{1}$\thanks{Equal contribution.}, Xinyu Zhang$^{1}$\footnotemark[1], Dong Wang$^{1}$\thanks{Corresponding Author: Dr. Dong Wang, wdice@dlut.edu.cn},  Huchuan Lu$^{1,2}$ and Xiaoyun Yang$^3$\\
$^1$School of Information and Communication Engineering, Dalian University of Technology, China\\
$^2$Peng Cheng Laboratory
$^3$Remark AI\\
{\tt\footnotesize \{yan\_bin,zhangxy71102\}@mail.dlut.edu.cn, 
\{wdice, lhchuan\}@dlut.edu.cn, xyang@remarkholdings.com}
}

% \and

\maketitle
\pagestyle{empty}
\thispagestyle{empty}

%%%%%%%%% ABSTRACT
\begin{abstract}
Visual object tracking aims to precisely estimate the bounding box for the given target, which is a challenging problem due to factors such as deformation and occlusion. Many recent trackers adopt the multiple-stage 
strategy to improve bounding box estimation. These methods first coarsely locate the target and then 
refine the initial prediction in the following stages. However, existing approaches still suffer from limited precision, 
and the coupling of different stages severely restricts the method’s transferability. This work proposes a novel, 
flexible, and accurate refinement module called Alpha-Refine (AR), which can significantly improve the base trackers’ 
box estimation quality. By exploring a series of design options, we conclude that the key to successful refinement is 
extracting and maintaining detailed spatial information as much as possible. Following this principle, Alpha-Refine 
adopts a pixel-wise correlation, a corner prediction head, and an auxiliary mask head as the core components. 
% >>>
\ignore{We apply Alpha-Refine to six famous base trackers to verify our method’s effectiveness: ECO, RT-MDNet, SiamRPNpp, ATOM, DiMP50, and DiMPsuper.}
Comprehensive experiments on TrackingNet, LaSOT, GOT-10K, and VOT2020 benchmarks with multiple base trackers show that our approach significantly improves the base tracker’s performance with little extra latency. 
% >>>
The proposed Alpha-Refine method leads to a series of strengthened trackers, among which the ARSiamRPN (AR strengthened SiamRPNpp) and the ARDiMP50 (AR strengthened DiMP50) achieve good efficiency-precision trade-off, while the ARDiMPsuper (AR strengthened DiMPsuper) achieves very competitive performance at a real-time speed.
Code and pretrained models are available at \href{https://github.com/MasterBin-IIAU/AlphaRefine} {https://github.com/MasterBin-IIAU/AlphaRefine}.
\end{abstract}

%%%%%%%%% BODY TEXT

% ############################## Introduction ################################
\section{Introduction}
% Introduction 1 page
% Visual object tracking requires the tracker not only to locate the given object (precision score), but also to estimate the bounding box of the object (AUC score). To this end, 
Precise box estimation is indispensable for a successful tracker.
Early trackers usually solve this problem by multi-scale search~\cite{SiameseFC,ECO,DRT,unveiling} 
or sampling-then-regression strategy~\cite{SINT,MDNet}, which are inaccurate and greatly limit the performance of the trackers. 
For obtaining more robust and precise tracking results, many state-of-the-art trackers~\cite{SPM,CascadedSiameseRPN,ATOM,DiMP} 
adopt a multiple-stage tracking strategy, which introduces additional tracking stages for more precise box estimation. 
These trackers first coarsely locate the target and then refine the initial result in the additional tracking stages 
to get more precise box prediction.
However, this box estimation can still be improved, even for state-of-the-art trackers.
An oracle experiment verifies this opinion. 
We set the tracker’s search region always centering at the ground truth so that
the performance will be mainly determined by the capacity of box estimation.
Table~\ref{tab:oracle} shows that with the aforementioned oracle setting, state-of-the-art trackers' 
AUC scores are still far from perfection, indicating unsatisfying box estimation, although 
some methods (\emph{e.g.} ATOM~\cite{ATOM}, DiMP~\cite{DiMP}) have built-in box estimation modules. 
In contrast, given the perfectly centered search region, the proposed Alpha-Refine achieves significantly 
better performance, demonstrating that Alpha-Refine's superiority in box estimation.

% <<<<<<<<<<<<<< Oracle Experiment <<<<<<<<<<<<<<<<<<<
\begin{table}[!t]
\caption{Oracle experiment on LaSOT. The center of the search region is always set at the center of the ground truth, 
reflecting the box estimation capacity of these methods. The best three results are marked in \textbf{\textcolor[rgb]{1,0,0}{red}}, \textbf{\textcolor[rgb]{0,1,0}{green}} and \textbf{\textcolor[rgb]{0,0,1}{blue}} bold fonts respectively.} 
\vspace{-1mm}
\begin{center}
\begin{tabular}{c|c|c|c}
\hline
\textbf{Oracle} & {\rm AUC} & ${\rm P_{Norm}}$ & {\rm P} \\
\hline
SiamRPNpp\cite{SiamRPNplus}&\textbf{\textcolor[rgb]{0,0,1}{0.682}}&\textbf{\textcolor[rgb]{0,1,0}{0.829}} & \textbf{\textcolor[rgb]{0,1,0}{0.745}}\\
ATOM\cite{ATOM}&0.580 &0.686 & 0.604  \\
DiMPsuper\cite{DiMP}&\textbf{\textcolor[rgb]{0,1,0}{0.693}}&\textbf{\textcolor[rgb]{0,0,1}{0.799}} & \textbf{\textcolor[rgb]{0,0,1}{0.734}}  \\
ECO\cite{ECO}&0.496&0.666 & 0.533  \\
% RTMDNet\cite{RTMDNet}&0.32&0.667&0.565  \\  % RTMDNet-oracle indeed not work
\hline
AlphaRefine&\textbf{\textcolor[rgb]{1,0,0}{0.762}}&\textbf{\textcolor[rgb]{1,0,0}{0.902}} & \textbf{\textcolor[rgb]{1,0,0}{0.919}}  \\
\hline
\end{tabular}
\end{center}
\label{tab:oracle}
\vspace{-6mm}
\end{table}

Additionally, most of refinement methods in existing trackers~\cite{SPM,CascadedSiameseRPN,ATOM,DiMP} are weak in transferability,
because their training is coupled with other components. Extra retraining is required if above refinement modules are applied to new base trackers.
As opposed to these methods, Alpha-Refine is trained independently and can be directly applied to any existing trackers in a plug-and-play style, requiring no extra training or modification of the base tracker. 

In this work, a series of design options are investigated and compared. Specifically, we assess multiple feature fusion modules and prediction heads. We also explore to use an auxiliary mask head, which introduces pixel-level supervision into the training. We find that extracting and maintaining precise spatial information is the key to precise box estimation.
To this end, we finally adopt a pixel-wise correlation as well as a key-point style prediction head for better maintaining and utilizing the detailed spatial information. Additionally, an auxiliary mask head is used, which encourages the network to extract more detailed spatial information, leading to more precise box estimation.
Moreover, if we reserve the mask head at the inference stage, Alpha-Refine will enable the base trackers to predict the mask of the object, satisfying scenarios where the mask is required.
The design options will be discussed in Section~\ref{sec:method} and verified in Section~\ref{sec:experiment}. 

\begin{figure}[!t]
    \begin{center}
        \includegraphics[width=1.0\linewidth]{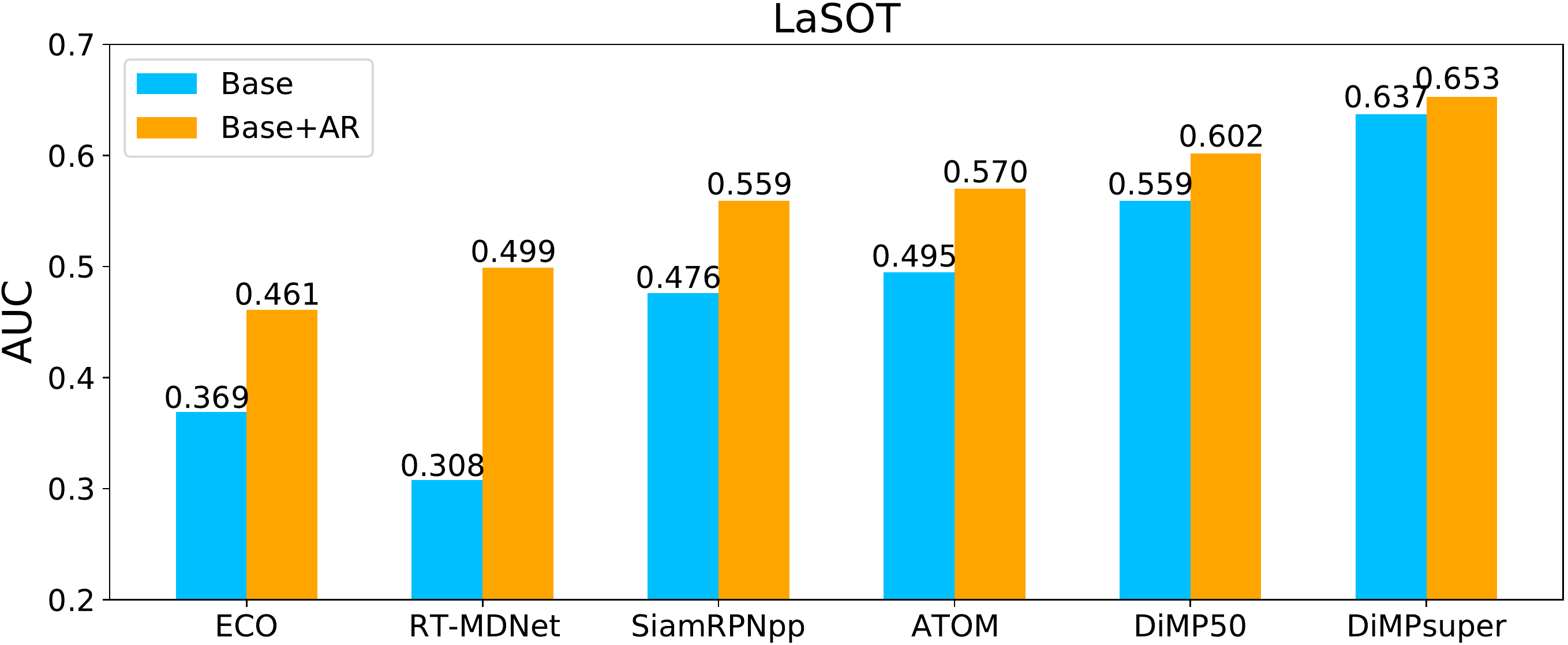}
        \end{center}
        \vspace{-2mm}
        \caption{Performance improvement of our Alpha-Refine module on LaSOT. 
        `Base': the base tracker; `Base+AR': the base tracker with Alpha-Refine (AR). This figure shows 
        that all base trackers are significantly improved by the proposed AR module.}
        \label{LaSOT_Improvement}
        \vspace{-2mm}
\end{figure}
To verify the effectiveness of our Alpha-Refine module, we choose six famous base trackers: ECO~\cite{ECO}, 
RT-MDNet~\cite{RTMDNet}, SiamRPNpp~\cite{SiamRPNplus}, ATOM~\cite{ATOM}, DiMP~\cite{DiMP}, and 
DiMPsuper~\cite{DiMP}, on multiple tracking benchmarks, namely, LaSOT~\cite{LaSOT}, GOT-10K~\cite{GOT10K}, TrackingNet~\cite{Trackingnet} and VOT2020~\cite{VOT2020}.
Take Fig.~\ref{LaSOT_Improvement} as an example, experimental results show that our proposed refinement module improves the base trackers' performance significantly. 
Compared with its competitors (\emph{i.e.} IoU-Net~\cite{ATOM,DiMP} and SiamMask~\cite{SiamMask}), Alpha-Refine's performance also surpasses them by a large margin. 

% >>>
The proposed Alpha-Refine method leads to a series of strengthened trackers, among which the ARSiamRPN (Alpha-Refine strengthened SiamRPNpp) and ARDiMP50 (Alpha-Refine strengthened DiMP50) achieve good efficiency-precision trade-off, while ARDiMPsuper (Alpha-Refine strengthened DiMPsuper) achieves state-of-the-art performance at a real-time speed on multiple benchmarks.

% ############################## Related Works ################################
\section{Related Works}
{\noindent \textbf{Early Box Estimation.}} Early box estimation methods are mainly scale estimation, 
which can be summarized into two categories:
multiple-scale search and sampling-then-regression strategies. 
Most correlation-filter-based trackers ~\cite{KCF,ECO,DRT} and SiamFC~\cite{SiameseFC} adopt the former strategy. 
Specifically, these trackers construct search regions with different sizes, then compute correlation with the template,
and finally determine the size of the target as the size-level where the highest response locates. 
Multiple-scale search is coarse and time-consuming due to its fixed-aspect-ratio prediction and heavy image pyramid 
operation. 
Another type of method first generates several bounding box samples, then uses some methods to choose 
the best one, and finally apply regression on it to obtain more accurate results. 
SINT~\cite{SINT}, MDNet~\cite{MDNet} and RT-MDNet~\cite{RTMDNet} are three representative trackers 
that exploit this approach. 

{\noindent \textbf{Modern Box Estimation.}} 
As deep learning techniques become mature, several high-performance scale 
estimation approaches are developed and can be categorized into the following classes: RPN-based~\cite{SiameseRPN,DSiam,SiamRPNplus}, Mask-based\cite{SiamMask,D3S}, IoU-based~\cite{ATOM,DiMP},
and Anchor-free-based~\cite{SiamFC++,TransT}. 
RPN-based methods learn a region proposal network~\cite{FasterRCNN}, which determines whether the current anchor 
contains the target and makes refinement to the target simultaneously. 
SiameseRPN-series trackers~\cite{SiameseRPN,DSiam,SiamRPNplus} utilize the RPN-based 
mechanism as the core component and achieve great success in recent years. 
Mask representation is more accurate, and the ability to predict mask is quite 
beneficial to precise box estimation.
SiamMask~\cite{SiamMask} and D3S~\cite{D3S} belong to this class, which obtain higher precision 
than the Siamese tracker that can only predict boxes.
IoU-based approaches learn a network to predict the overlap between candidate boxes and groundtruth. 
During the inference phase, this strategy optimizes candidate boxes by gradient-ascent, 
and therefore obtains more precise results. 
ATOM~\cite{ATOM} and DiMP~\cite{DiMP} fully exploit this method and surpass traditional 
correlation-filter trackers by a large margin. 
In the past years, anchor-free philosophy has become quite popular in the object detection 
field~\cite{CornerNet,CenterNet,Foveabox,FCOS}. 
SiamFC++~\cite{SiamFC++} introduces this structure into object tracking field and therefore 
achieves state-of-the-art performance. 
The CGACD method~\cite{CGACD} designs a corner-based box estimation for object tracking which wisely adopts soft-argmax to decode the corner heat map into box coordinates. The corner-based version of Alpha-Refine 
adopts a similar box representation, and experiments demonstrate that this design retains more precise 
spatial information for a refinement module.

{\noindent \textbf{Refinement Modules.}} 
Many state-of-the-art trackers~\cite{SPM,CascadedSiameseRPN,ATOM,DiMP,VOT2019} apply a multiple-stage 
tracking strategy to obtain accurate and robust results. 
This approach first locates the target coarsely and then utilizes a refinement module to refine results from 
the previous stage. 
SPM~\cite{SPM} and Siamese Cascaded RPN~\cite{CascadedSiameseRPN} adopt a light-weight relation 
network~\cite{RelationNet} and stacked RPNs~\cite{FasterRCNN}, respectively, as the refinement module 
to further increase trackers' discriminative power and precision. 
However, the two refinement modules have to be trained together with their previous Siamese tracker 
in an end-to-end manner; this procedure limits their flexibility of combining with other base trackers. 
ATOM~\cite{ATOM} and DiMP~\cite{DiMP} first use an online classification module to locate the target 
and then draw some random samples around it. 
Finally, they deploy a modified IoU-Net~\cite{IOU-Net} to 
maximize the overlap between these samples and groundtruth to obtain more precise bounding boxes. 
This modified IoU-Net can be trained separately from the base tracker. 
Thus, the IoU-Net has good transferability but its precision can still be greatly improved. 
Notably, the winners of VOT2019~\cite{VOT2019} utilize 
SiamMask~\cite{SiamMask} as a refinement module~\cite{VOT2019}. 
Similar to IoU-Net~\cite{IOU-Net} mentioned before, SiamMask~\cite{SiamMask} can be combined 
with any base tracker. 
However, SiamMask is designed as an independent tracker rather than a refinement module, 
which is not suitable and not economical to refine other trackers. 
Considering previous refinement modules' weak transferability and limited accuracy, 
we propose a novel, flexible, and accurate refinement module named Alpha-Refine. 

\begin{figure*}[t]
    \begin{center}
    \includegraphics[width=1.0\linewidth]{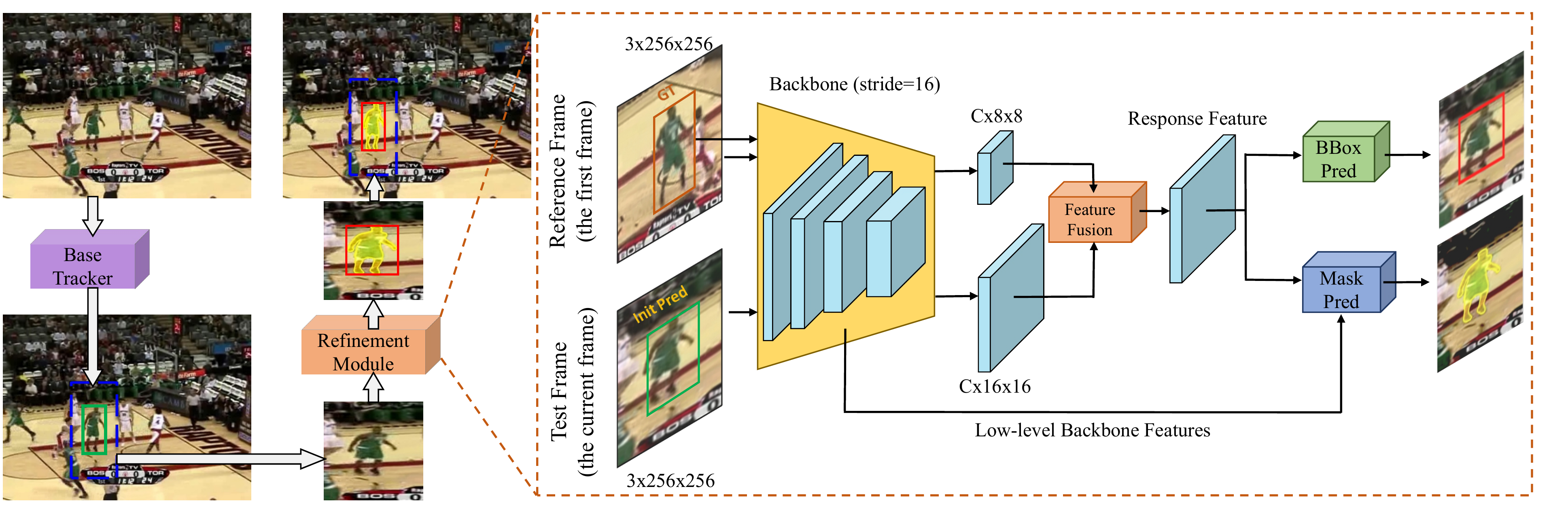}
    \end{center}
    \vspace{-1mm}
    \caption{Overall architecture of the proposed Alpha-Refine. Better viewed in color with zoom-in.}
    \label{fig-architecture}
\end{figure*}

% ############################## Our Method ################################
\section{Alpha-Refine}\label{sec:method}
Alpha-Refine is a refinement module which is able to efficiently refine the base tracker's outputs and significantly improve the tracking performance. 
We detail the network architecture (Fig.~\ref{fig-architecture}), design options, and training process as follows. 

\subsection{Network Architecture} 
Fig.~\ref{fig-architecture} shows the overall architecture of the proposed Alpha-Refine module. 
This module adopts the Siamese architecture with two input branches, namely, the \textit{reference branch} 
and the \textit{test branch}. 
A parameter-shared backbone is applied to both branches for feature extraction. Features extracted from 
two branches are aggregated by a fusion module, which is typically a correlation module (\emph{e.g.} 
naive-correlation, depth-wise correlation, pixel-wise correlation). 

The fused feature is further processed by some convolutional layers, producing the features for the prediction head. 
An auxiliary mask head can be added parallel to the box head to introduce pixel-level supervision into training. 
The output of the mask head can be used for scenarios that also require a mask result.

To function as a refinement module, the reference branch is initialized by the first 
frame with the ground truth. 
In the current frame, the test branch extends the base tracker’s prediction into a concentric search region  
of two times the size, from which Alpha-Refine predicts a finer result. 
Alpha-Refine can be combined with arbitrary trackers in a plug-and-play style and improve their performance.

Notably, compared with independent trackers, the size of Alpha-Refine’s search region is roughly two times the size
of the object, which is smaller than normal trackers (four times in most cases). 
The smaller search region can depress the cluttered background and enable the model to focus on more detailed spatial
information, which is beneficial to precise localization.
Small search region also lowers the computation cost, so that Alpha-Refine can improve the base tracker with little
latency increase. 
Alpha-Refine is not capable of tracking by itself because of the small search region light-weight 
design. A complete base tracker is always needed.

\subsection{Feature Fusion}\label{sec:featfuse}
\begin{figure}
\centering
    \includegraphics[width=0.7\linewidth]{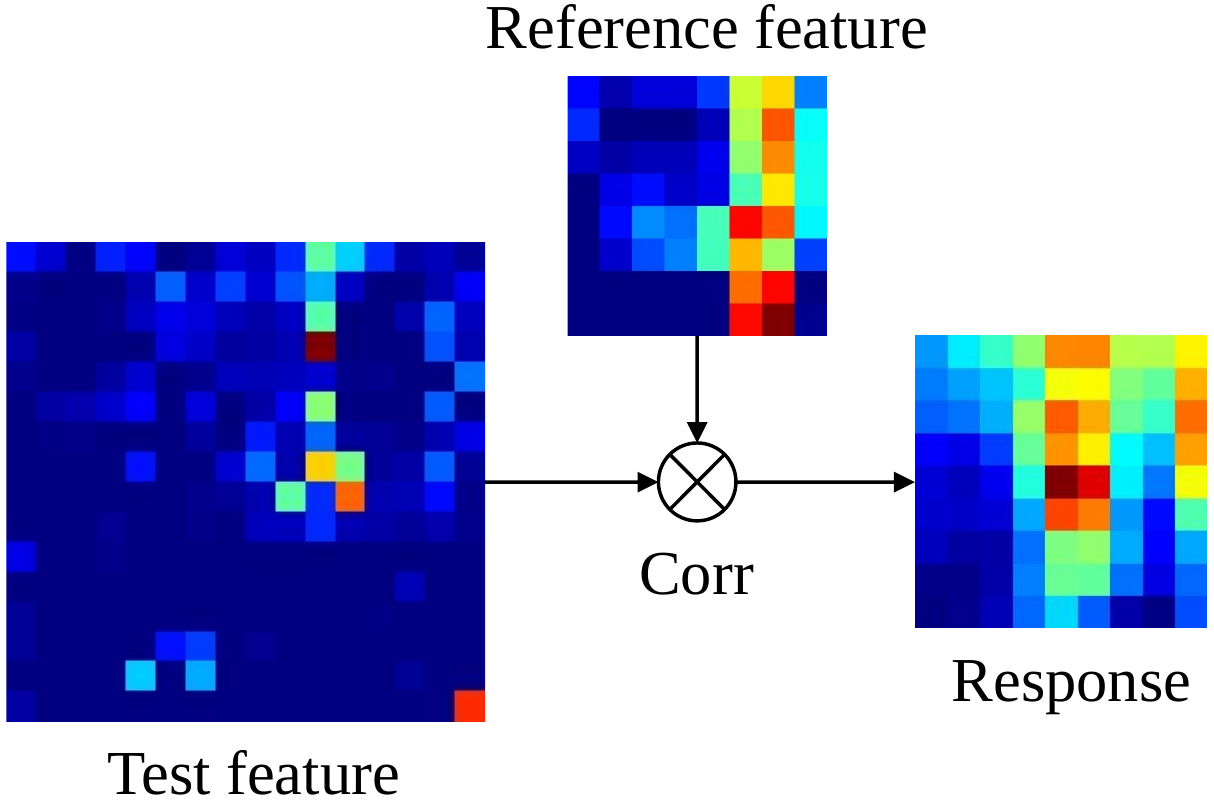}
    \caption{Illustration of the blur effect. Depth-wise correlation or naive correlation may blur the spatial information.} \label{fig:dwcorr_blur}
    \vspace{-3mm}
\end{figure}

Most methods with Siamese architecture aggregate features of the template and the search region using the coarse naive correlation~\cite{SiameseFC,SiameseRPN,DSiam} or depth-wise correlation~\cite{SiamRPNplus,SiamMask}. 
As shown in Fig.~\ref{fig:dwcorr_blur}, both naive correlation or depth-wise correlation take the whole template feature 
as the kernel to correlate with the search region feature, making adjacent sliding window on the feature map producing 
similar response and blur the spatial information. 
As a refinement module, Alpha-Refine requires the feature fusion module to maintain as much spatial information as possible.
Thus, the popular naive nor depth-wise correlation is not suitable for Alpha-Refine.

%Fig.~\ref{fig:dwcorr_blur} shows that naive and depth-wise correlations take the entire reference feature (feature from 
%reference branch) as the kernel to correlate with the test feature (feature from test branch), which enables the adjacent 
%sliding window on the test feature to produce a similar response.
%Thus, when the two features correlate with each other, the location information will be blurred. 

In this work, we adopt pixel-wise correlation~\cite{RANet} for high-quality feature representation. 
We denote $K \in {\mathbb{R}^{C \times {H_0} \times {W_0}}}$ and $S \in {\mathbb{R}^{C \times H \times W}}$ 
as features of the template and the search region. 
Pixel-wise correlation first decomposes $K$ into ${H_0}{W_0}$ small kernels 
${K_j} \in {\mathbb{R}^{C \times {1} \times {1}}}$ and then uses them to compute correlation 
separately to obtain correlation maps $C \in {\mathbb{R}^{{H_0}{W_0} \times H \times W}}$. 
The process can be described as
\begin{equation}
C = {\{ {C_j}|{C_j} = {K_j}*S\} _{j \in \{ 1,...,{H_0} \times {W_0}\} }},
\end{equation}
where $*$ denotes naive correlation.

In contrast to naive or depth-wise correlation, pixel-wise correlation takes each part of the target features as a kernel. 
Pixel-wise correlation ensures that each correlation map encodes information of a local region on the target 
while avoiding an extremely large correlation window from blurring the feature. 

\begin{figure*}
    \includegraphics[width=1.0\linewidth]{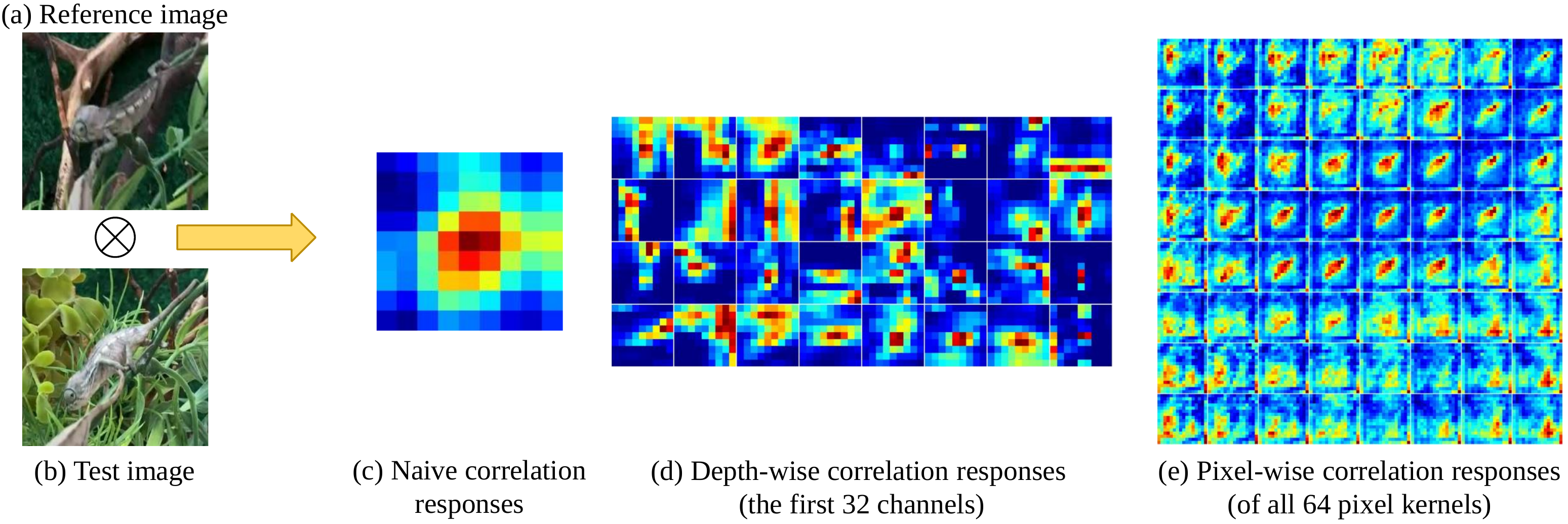}
    \caption{Comparison among different correlation responses. (a) and (b) denote the reference branch and 
    the test branch, respectively. (c), (d), and (e) are correlation result of naive, depth-wise (the first 32 channels of 256), and pixel-wise correlations, respectively.} \label{fig:featfuse}
\end{figure*}

\begin{figure}
    \includegraphics[width=1.0\linewidth]{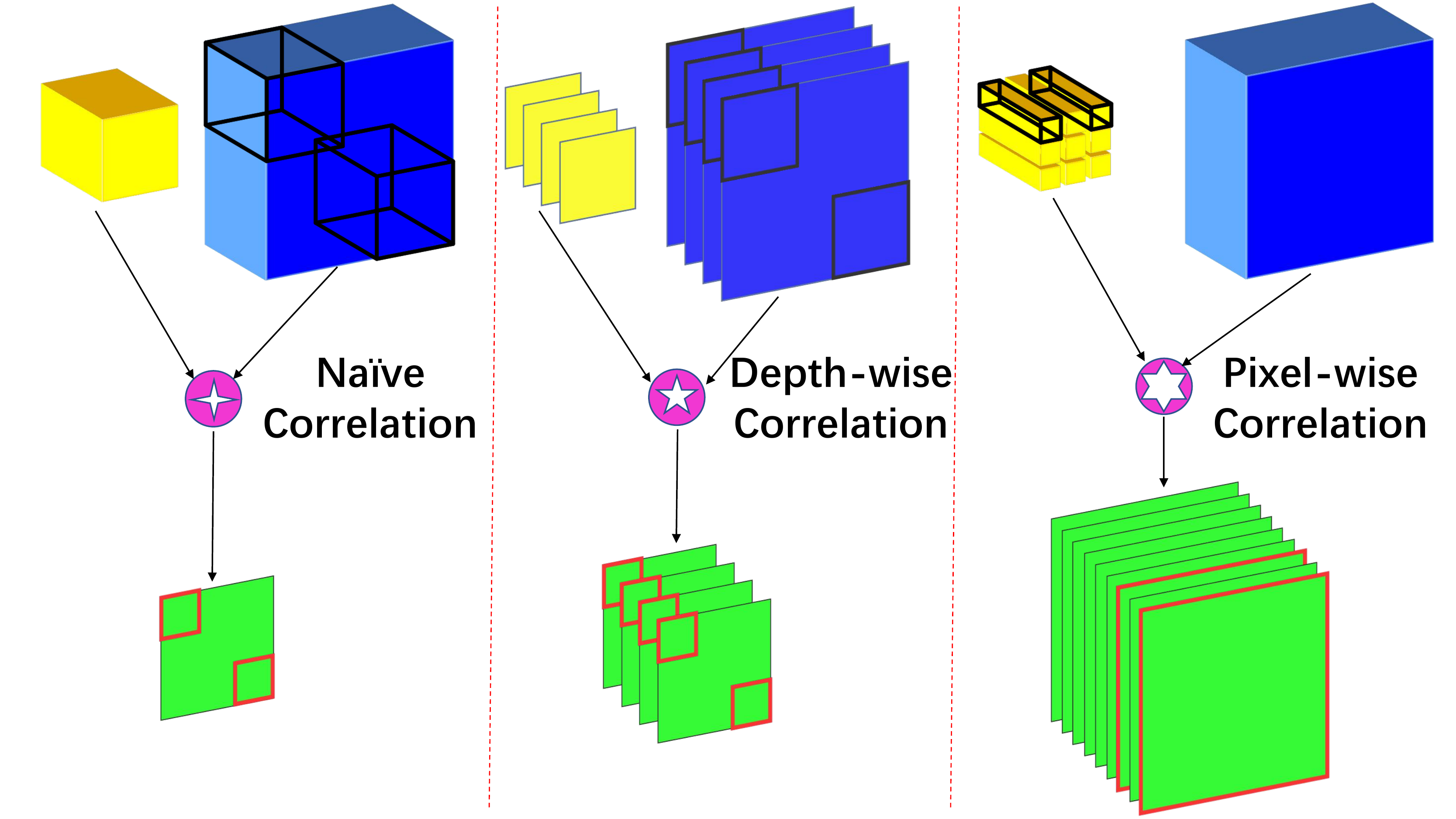}
    \caption{Comparison among different correlation methods. From left to right, naive,
depth-wise, and pixel-wise correlations are demonstrated.
The black-edged cubes or squares represent sliding kernels. The
red edged ones represent corresponding correlation maps.} \label{fig-corr-NL}
\end{figure}

Fig.~\ref{fig-corr-NL} shows the computation process of three correlation methods, 
and Fig.~\ref{fig:featfuse} shows some fusion outputs.
Fig.~\ref{fig:featfuse}(c) indicates that naive convolution can only roughly represent the center location 
of the object while losing most of the shape and scale information. 
Fig.~\ref{fig:featfuse}(d) illustrates that depth-wise correlation has to encode the blurred location into channels, 
which is less explainable and inefficient. 
By contrast, Fig.~\ref{fig:featfuse}(e) shows that pixel-wise correlation is better at retaining the target's 
boundary and other detailed spatial information. 

\subsection{Prediction Heads}
We explore two ways of predicting the bounding box: directly regressing the box coordinates 
and predicting two corner points\footnote{The top-left corner and the bottom-right corner} 
from two heatmaps. For regressing the box coordinates, we evaluate RPN style and RCNN 
style designs. For predicting the corners, we evaluate the key-point style design.

\begin{figure}[!htb]
\centering
    \includegraphics[width=1\linewidth]{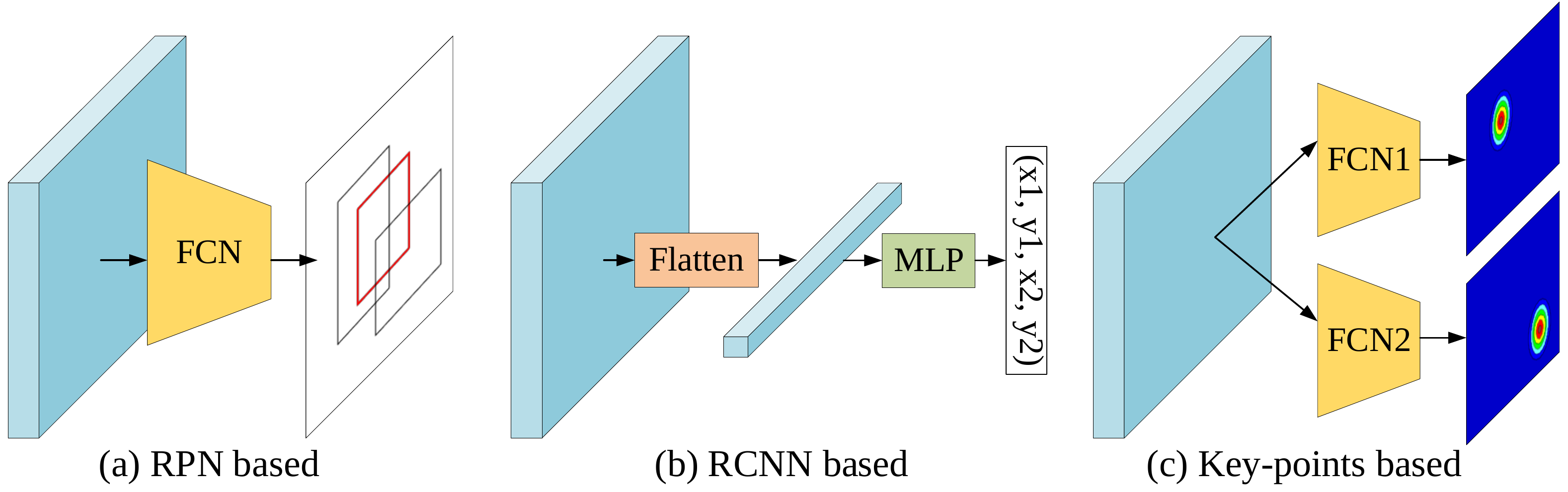}
    \caption{Options for predicting box results. (a) RPN style box prediction. (b) RCNN style box prediction. (c) Corner prediction.} \label{fig:boxhead}
    \vspace{-4mm}
\end{figure}

\noindent{\textbf{RPN Style Box Head. }}
One way of regressing the bounding box is the RPN Style Box Head. 
Fig.~\ref{fig:boxhead}(a) shows the diagram.
Similar to the FCN structure of one-stage object detectors, a 4D box coordinates 
together with a confidence score is predicted at each location. 
The box with the highest score is regarded as the tracking result. 

However, we notice some drawbacks of using this method in the refinement module.
%
%In this method, each box prediction is generated by an individual feature point, 
%which requires this feature point to encode the spatial information within its receptive field into the channels. 
%Considering the translation equivariance of network, the 2D structure of the feature map contains rich spatial information, which is not fully %utilized by individual feature points. Additionally, inconsistency exists in this strategy, because the most precise box prediction may get a low %confidence. 
In this method, each box prediction is generated by an individual feature point, 
which requires this feature point to summarize the information in its receptive field and encode spatial
information into the channels, so that a single feature point can make a prediction by itself.
However, different feature points have varying receptive fields, making them good at representing various parts of the object. 
The RPN style method ignores the relationship between feature points at different locations, not fully utilizing the information
contained in the spatial distribution of the feature map. Additionally, this strategy has inconsistency because most
precise box predictions may have a low score.
In the experiment, we implement this strategy by stacking four Conv-BN-ReLU layers followed by a prediction layer. 
For simplicity, we directly regress four distances from the feature point location to four edges of the bounding box. 
Another four Conv-BN-ReLU layers are used to predict the confidence score. 

\noindent{\textbf{RCNN Style Box Head. }}
Similar to the second stage of Faster-RCNN~\cite{FasterRCNN}, this RCNN Style method reduces the feature map into a vector and 
estimates the bounding box of the object with some fully connected layers. 
Fig.~\ref{fig:boxhead}(b) shows the diagram. 
Compared with the RPN style method, this method utilize the whole feature map rather than individual feature points. 
%
%However, the 2D structure of the feature map is not retrained. The network still encode the spatial information 
%into channels, where the loss of detail is unavoidable. 
Apparently, this method crashes spatial information when reducing the feature map, indicating that it is not suitable
for a refinement module.
For this strategy, we use a bounding box head containing four stacked Conv-BN-ReLU layers in our experiment, 
followed by a global average pooling layer and a fully-connected layer, which predicts four coordinates of the bounding box. 

\noindent{\textbf{Corner Head. }} Recently, keypoints detection techniques have become popular in the object detection field, 
producing several state-of-the-art methods~\cite{CornerNet,CenterNet,centernet-triplet,extremenet}. 
In our experiment, we implement a corner head with four stacked Conv-BN-ReLU layers, 
followed by a Conv layer predicting two heatmaps, which represent top-left corner and bottom-right corner respectively. 
Different from methods like CornerNet~\cite{CornerNet}, we do not upsample the feature map for the computation issue, 
resulting in a coarse-grained heatmaps.
We apply soft-argmax~\cite{soft-argmax} to the heatmaps to make the discrete heatmaps precisely describe the position of the corner point.
%derive the mass centers of both heatmaps. The mass centers represent the top-left and bottom-right corner respectively.
%
The soft-argmax operation enable our model to predict continuous values from discrete heatmaps. It encodes the box estimation into the distribution of confidence (heat) scores, avoiding the inconsistency problem in the RPN style head. The key-point style method retrains the natural spatial structure of the feature map, avoiding encoding spatial information into channels, which is desirable for Alpha-Refine. 

\noindent{\textbf{Auxiliary Mask Head. }} 
As Alpha-Refine is a module for precisely estimating the bounding box, additional detailed shape information would be helpful. 
To this end, we add an auxiliary mask head parallel to the box head, which introduces pixel-level supervision into training.
When the box head is trained with the auxiliary mask head, the network is encouraged to extract more detailed spatial information 
which is required by the mask head and facilitates precise box estimation. 
In addition, supervision from mask annotation also teach the model to better discriminate foreground and background, 
which is required by the segmentation task and also beneficial to tracking.
Some previous works~\cite{SiamMask,D3S} also demonstrate that mask prediction is quite beneficial for improving tracking 
performance, especially on benchmarks (\emph{e.g.}, VOT~\cite{VOT2018,VOT2019}) that adopt rotated bounding box labels. 
In this work, the mask head is implemented as a U-Net~\cite{U-Net} style decoder, which gradually upsample the feature 
map while fusing them with low-level features from the backbone until the resolution is the same as the input image, 
and a mask is predicted from the last layer. At the inference stage, the mask head is by default disabled to speed up 
Alpha-Refine. For scenarios requiring pixel-level prediction, the mask head can be activated, producing mask prediction as the output.
%
% Different from SiamMask~\cite{SiamMask} that restricts the predicted mask in a region as large as the template, our mask branch predicts mask that has the same size as the search region for a higher-quality mask output. 

\subsection{Training}
{\flushleft \textbf{Training Set Construction.}} We use the training splits of LaSOT~\cite{LaSOT} and 
GOT-10K~\cite{GOT10K}, ImageNet VID~\cite{ImageNet}, ImageNet DET~\cite{ImageNet}, COCO~\cite{COCO}, 
Youtube-VOS~\cite{youtube-vos}, and some segmentation datasets~\cite{DUT-OMRON,DUTS,ECSSD} 
to train the Alpha-Refine. 
Given a video sequence, two stochastic frames $F_{ref}$ and $F_{test}$ with an interval of less than 
$50$ frames are first selected. 
The input of the reference branch is obtained by cropping $F_{ref}$ at the center of the ground truth 
with two times the size of the ground truth box. 
The input of the test branch is obtained by cropping $F_{test}$ randomly centered around the ground truth, 
with a jittered size. 
Specifically, we randomly translate and scale the ground truth of $F_{test}$ to obtain the region 
to be cropped.
With the following equations:

\begin{equation}
[h,w] = [{2h^{GT}},{2w^{GT}}] \times {e^{{\bf{N}}{f_s^{test}}}}
\end{equation}
\begin{equation}
O{_{\max }} = \sqrt {hw}  \times {f_c^{test}}
\end{equation}
\begin{equation}
[{c_x},{c_y}] = [c_x^{GT},c_y^{GT}] + ({\bf{U}} - 0.5) \times O{_{\max }}
\end{equation}
we obtain the region centering at $[{c_x},{c_y}]$ with size $[h,w]$. 
$[c_x^{GT},c_y^{GT},{h^{GT}},{w^{GT}}]$ is the ground truth bounding box.
$f_s^{test}$ and $f_c^{test}$ are two scalar factors corresponding to scale and center, respectively. 
We use $[f_s^{test},f_c^{test}] = [0.25,0.25]$ in our experiments. 
$\bf{N}$ and $\bf{U}$ represent the $2D$ standard normal distributed random variable and $2D$ 
uniform random variable respectively. 
The cropped images are resize into $256 \times 256$ as the inputs of Alpha-Refine. 

{\noindent \textbf{Training Approach.}} 
For the box output (\emph{i.e.} output of RPN style, RCNN syle, Key-Point style Heads), 
the mean squared error ${L_{box}}$ is used. 
All predictions are converted into coordinate vectors of the format 
[left-most, top-most, right-most, bottom-most] 
and compared with ground truth to obtain the mean squared error.
For the mask output, a binary cross-entropy loss ${L_{mask}}$ is used.
The total loss $L$ is the weighted sum of two losses.
\begin{equation}
L = {L_{box}} + {\lambda}{L_{mask}},
\end{equation}
where ${\lambda} = 1000$ is used in the experiments. 
We train Alpha-Refine for $40$ epochs, each of which consists of 500 iterations 
on eight Nvidia 2080Ti GPU with a batch size of 32 per GPU ($32\times8$ samples 
per iteration in total). 
Considering the abundance of the training data, we do not freeze any parameters of the backbone. 
The Adam optimizer~\cite{Adam} is applied and the learning rate halves every $8$ epochs. 

% ############################## Experiments ################################
\section{Experiments}\label{sec:experiment}
We implement our algorithm with the Pytorch~\cite{pytorch} deep learning library.
%, and the source codes are available in the supplementary material.
%
In this section, we verify the effectiveness of Alpha-Refine by performing comprehensive 
experiments on many popular tracking benchmarks: LaSOT~\cite{LaSOT}, TrackingNet~\cite{Trackingnet}, GOT-10K~\cite{GOT10K}, 
and VOT2020~\cite{VOT2020} together with six representative and state-of-the-art base trackers (including ECO~\cite{ECO}, 
RT-MDNet~\cite{RTMDNet}, ATOM~\cite{ATOM}, SiamRPNpp~\cite{SiamRPNplus}, DiMP50~\cite{DiMP}, and DiMPsuper~\cite{DiMP}) 
to demonstrate our Alpha-Refine's capacity of boosting the trackers' performance. 
Besides, we evaluate our design options with SiamRPNpp~\cite{SiamRPNplus} as the base tracker and 
determine the effects of different settings of our Alpha-Refine.
ResNet-34 is used as backbone by default if not otherwise specified. 
All experiments of our trackers run five times, and the results are obtained by average. 

\subsection{Representative Visual Results}
\begin{figure}[!h]
\centering
    \includegraphics[width=1.0\linewidth]{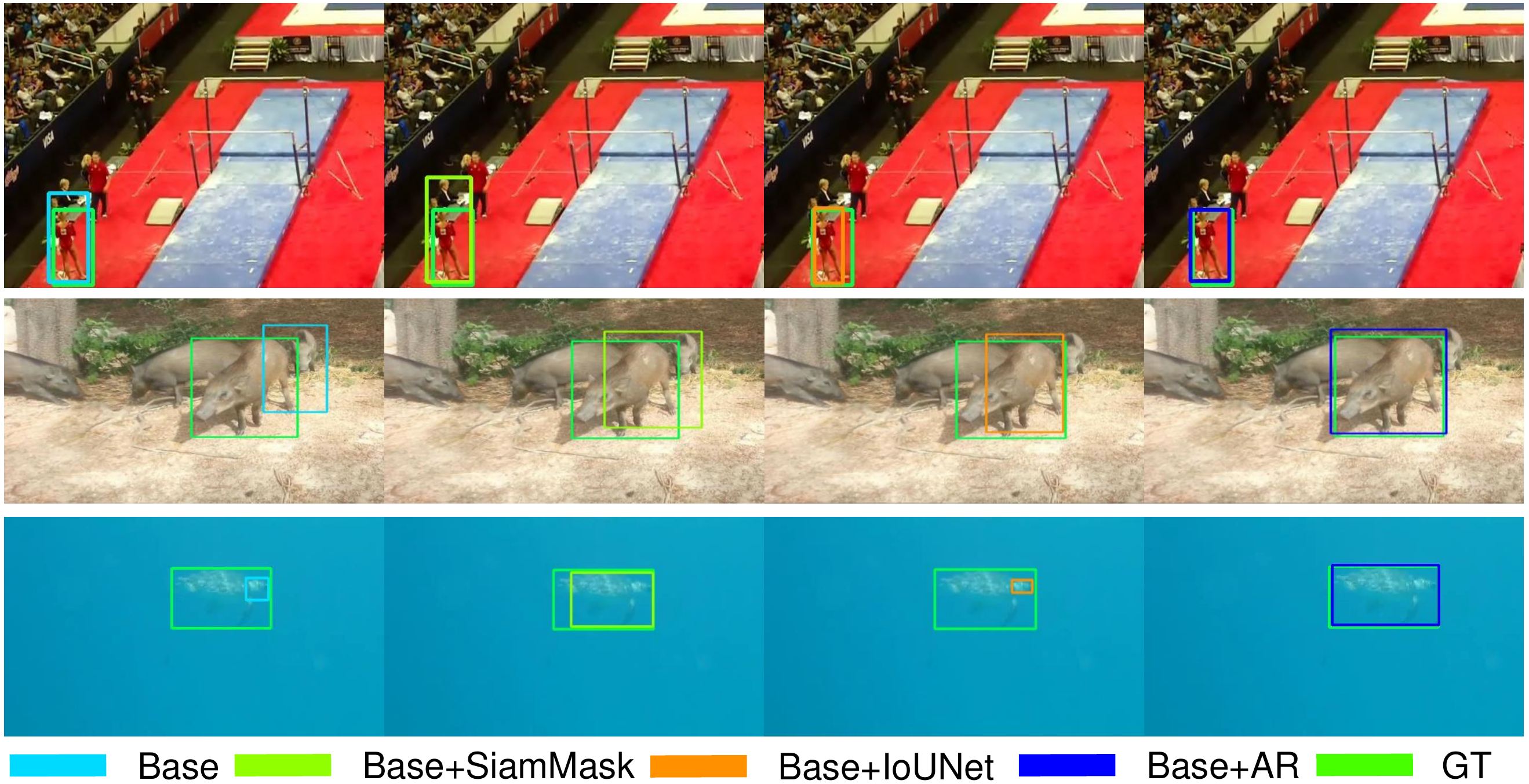}
%\caption{Visual Comparisons of Alpha-Refine and other refinement modules. From left to right, we present the \textcolor[rgb]{0, 0.85714275, 1}{origin prediction of base tracker}, and refined result by  \textcolor[rgb]{0.57142863, 1, 0}{SiamMask}, \textcolor[rgb]{1, 0.57142863, 0}{IoU-Net},  \textcolor[rgb]{0, 0, 1}{Alpha-Refine}. The \textcolor[rgb]{0.28571412, 1, 0}{ground truth} is presented for reference.} \label{fig:visual}
    \caption{Visual Comparison of Alpha-Refine and other refinement modules. From left to right, we present the origin prediction of the SiamRPNpp base tracker, and refined results obtained by SiamMask~\cite{SiamMask}, IoU-Net~\cite{ATOM,DiMP}, our Alpha-Refine.} \label{fig:visual}
    \vspace{-4mm}
\end{figure}

Figure~\ref{fig:visual} provides some representative visual results regarding different refinement 
module. We can see that our Alpha-Refine module facilitates the tracker obtaining more precise bounding 
boxes than IoU-Net and SiamMask.   
\vspace{-2mm}
%\emph{More visual results are presented in the supplementary material.} 

% \textcolor{red}{We need to provide legend for this figure!!}

\subsection{Evaluation on LaSOT}
\label{sec_lasot}
LaSOT~\cite{LaSOT} is a recent large-scale tracking benchmark, 
which consists of $1400$ challenging videos (1120 for training and 280 for testing). 
In this work, we follow the one-pass evaluation, using Success ($\rm{AUC}$), Normalized Precision 
($\rm{P_{Norm}}$), and Precision ($\rm{P}$), to compare different trackers without and with Alpha-Refine. 
Table~\ref{tab:lasot} shows that our Alpha-Refine (AR) module consistently and significantly 
improves the base trackers in all evaluation metrics. 
Especially for RT-MDNet, the improvement of the AUC score is up to $19\%$.
As shown in Table~\ref{tab:lasot-sota} the previous best tracker is Siam R-CNN~\cite{SiamRCNN}, which obtains a $64.8\%$ 
$\rm{AUC}$ score but merely runs around 5 \emph{fps}. 
in contrast, ARDiMPsuper achieves the best record (\rm{AUC}: 65.3\%), while maintaining a real-time speed. 

% Table: Comparison on LaSOT
\begin{table}[!t]
    \caption{Comparison results on the \textit{LaSOT test} set. 
    `Base': the base tracker; and `Base+AR': the base tracker with Alpha-Refine. 
    The best three results are marked in \textbf{\textcolor[rgb]{1,0,0}{red}}, \textbf{\textcolor[rgb]{0,1,0}{green}} and \textbf{\textcolor[rgb]{0,0,1}{blue}} bold fonts, 
    respectively. Numbers are shown in percentage (\%).}
    \centering
    \resizebox{\linewidth}{!}{
    \begin{tabular}{c|ccc|ccc}
        \hline
            \multirow{2}{*}{\textbf{Method}}&\multicolumn{3}{c|}{\textbf{Base}} &\multicolumn{3}{c}{\textbf{Base+AR}} \\
            \cline{2-7}
            &\rm{AUC}&$\rm{P_{Norm}}$&$\rm{P}$ &\rm{AUC}&$\rm{P_{Norm}}$&$\rm{P}$ \\
            \hline
        ECO         &36.9  &43.5  &36.4      &46.1  &50.8  &46.0 \\
        RT-MDNet    &30.8  &36.0  &30.1      &49.9  &63.1  &50.7 \\
        SiamRPNpp   &47.6  &54.7  &47.2      &55.9  &62.2  &57.4 \\
        ATOM        &\textbf{\textcolor[rgb]{0,0,1}{49.5}}  &\textbf{\textcolor[rgb]{0,0,1}{56.0}}  &\textbf{\textcolor[rgb]{0,0,1}{49.1}}      &\textbf{\textcolor[rgb]{0,0,1}{57.0}}  &\textbf{\textcolor[rgb]{0,0,1}{63.0}}  &\textbf{\textcolor[rgb]{0,0,1}{58.1}} \\
        DiMP50      &\textbf{\textcolor[rgb]{0,1,0}{55.9}}  &\textbf{\textcolor[rgb]{0,1,0}{63.3}}  &\textbf{\textcolor[rgb]{0,1,0}{55.3}}      &\textbf{\textcolor[rgb]{0,1,0}{60.2}}  &\textbf{\textcolor[rgb]{0,1,0}{66.8}}  &\textbf{\textcolor[rgb]{0,1,0}{61.7}} \\
        DiMPsuper   &\textbf{\textcolor[rgb]{1,0,0}{63.7}}  &\textbf{\textcolor[rgb]{1,0,0}{72.5}}  &\textbf{\textcolor[rgb]{1,0,0}{65.6}}     &\textbf{\textcolor[rgb]{1,0,0}{65.3}}   &\textbf{\textcolor[rgb]{1,0,0}{73.2}}  &\textbf{\textcolor[rgb]{1,0,0}{68.0}} \\
        \hline
    \end{tabular}}
    \label{tab:lasot}
    \vspace{-3mm}

\end{table}

% >>>
% Table: Benchmark on LaSOT
% \begin{table*}[!t]
%     \caption{Comparison state-of-the-art results on the \textit{LaSOT test} set. 
%     The best three results are marked in \textbf{\textcolor[rgb]{1,0,0}{red}}, \textbf{\textcolor[rgb]{0,1,0}{green}} and \textbf{\textcolor[rgb]{0,0,1}{blue}} bold fonts, 
%     respectively. Numbers are shown in percentage (\%).}
    
%     \centering
%     \resizebox{\linewidth}{!}{
%     \begin{tabular}{c|ccccccccc}
%         \hline
%         Method      &\textbf{ARDiMPsuper (ours)}  &SiamRCNN\cite{SiamRCNN}  &\textbf{ARDiMP50 (ours)}  &PrDiMP\cite{PrDiMP}  &LTMU\cite{LTMU} &DiMP50\cite{DiMP} &Ocean\cite{Ocean} &\textbf{ARSiamRPN (ours)}\\
%         \hline
%         AUC(\%)     &\textbf{\textcolor[rgb]{1,0,0}{65.3}}  &\textbf{\textcolor[rgb]{0,1,0}{64.8}}  &\textbf{\textcolor[rgb]{0,0,1}{60.2}} &59.8  &57.2  &56.8 &56.0  &56.0\\
%         Speed(fps)  &33  &5     &46 &30    &13    &59   &25    &50\\
%         \hline
%     \end{tabular}}
%     \label{tab:lasot-sota}
% \end{table*}

\begin{table*}[!t]
    \caption{Comparison state-of-the-art results on the \textit{LaSOT test} set. 
    The best three results are marked in \textbf{\textcolor[rgb]{1,0,0}{red}}, \textbf{\textcolor[rgb]{0,1,0}{green}} and \textbf{\textcolor[rgb]{0,0,1}{blue}} bold fonts, 
    respectively. Numbers are shown in percentage (\%). More results are available at \href{https://github.com/MasterBin-IIAU/AlphaRefine} {https://github.com/MasterBin-IIAU/AlphaRefine}.}
    
    \centering
    \resizebox{\linewidth}{!}{
    \begin{tabular}{c|cccccccccc}
        \hline
        \tabincell{c}{Method\\\quad}&\tabincell{c}{\textbf{ARDiMPsuper}\\(ours)}&\tabincell{c}{SiamRCNN\\\cite{SiamRCNN}}&\tabincell{c}{\textbf{ARDiMP50}\\ (ours)}  &\tabincell{c}{PrDiMP\\\cite{PrDiMP}}  &\tabincell{c}{LTMU\\\cite{LTMU}} &\tabincell{c}{DiMP50\\\cite{DiMP}} &\tabincell{c}{Ocean\\\cite{Ocean}} &\tabincell{c}{\textbf{ARSiamRPN}\\ (ours)}  &\tabincell{c}{SiamAttn\\\cite{SiamAttn}}  &\tabincell{c}{SiamFC++\\\cite{SiamFC++}} \\
        
        \hline
        AUC(\%)     &\textbf{\textcolor[rgb]{1,0,0}{65.3}}  &\textbf{\textcolor[rgb]{0,1,0}{64.8}}  &\textbf{\textcolor[rgb]{0,0,1}{60.2}} &59.8  &57.2  &56.8 &56.0  &56.0  &56.0  &54.4\\
        Speed(fps)  &33  &5     &46 &30    &13    &59   &25    &50  &45  &90\\
        \hline
    \end{tabular}}
    \label{tab:lasot-sota}
\end{table*}

\noindent{\textbf{Latency and Speed. }} 
Table~\ref{Ablation_Speed} reports the latency and speed performance of different trackers 
without and with Alpha-Refine, showing that our Alpha-Refine module introduces 
few computation loads (merely about 5-6ms every frame), while significantly improving the tracking 
accuracies (see Table~\ref{tab:lasot}).
\ignore{Note that the best tracker, ARDiMPsuper, also obtains real-time performance. }

\begin{table}[htbp]
\caption{Latency and speed of different methods. The tracking speed is measured using frame per second (fps).\label{Ablation_Speed}}
\vspace{-6mm}
\begin{center}
\resizebox{\linewidth}{!}{
\begin{tabular}{l|c|c|c|c|c}
\hline
\multirow{2}{*}{\textbf{Method}}  &\multicolumn{2}{c|}{\textbf{Base}}   &\multicolumn{2}{c|}{\textbf{Base+AR}}&\multirow{2}{*}{$\Delta t$}\\
\cline{2-5}
    	    &latency & fps 	&latency &fps 	 \\
\hline
ECO	        &13.3ms	& 75.2 	&18.9ms	& 52.9 	&+5.6ms \\
RTMDNet	    &14.3ms	& 69.9 	&20.1ms	& 49.8 	&+5.7ms \\
ATOM	    &16.8ms	& 59.5 	&22.1ms	& 45.2 	&+5.3ms \\
SiamRPNpp	&14.9ms	& 67.1 	&20.0ms	& 50.0 	&+5.1ms \\
DiMP50	    &16.7ms	& 59.9 	&21.9ms	& 45.7 	&+5.2ms \\
DiMPsuper	&25.2ms	& 39.7 	&30.4ms	& 32.9 	&+5.2ms \\
\hline
\end{tabular}}
\end{center}
\vspace{-6mm}
\end{table}

\ignore{
\noindent{\textbf{Attribute-based Evaluation.}} 
Figure~\ref{fig:lasotatt} provides an attribute-based evaluation of representative 
base trackers with and without Alpha-Refine, illustrating that the proposed AR module 
significantly improves the base trackers on almost all attributes. 
\emph{More comparison results are presented in the supplementary material.}

\begin{figure}[h]
\centering
    \includegraphics[width=0.9\linewidth]{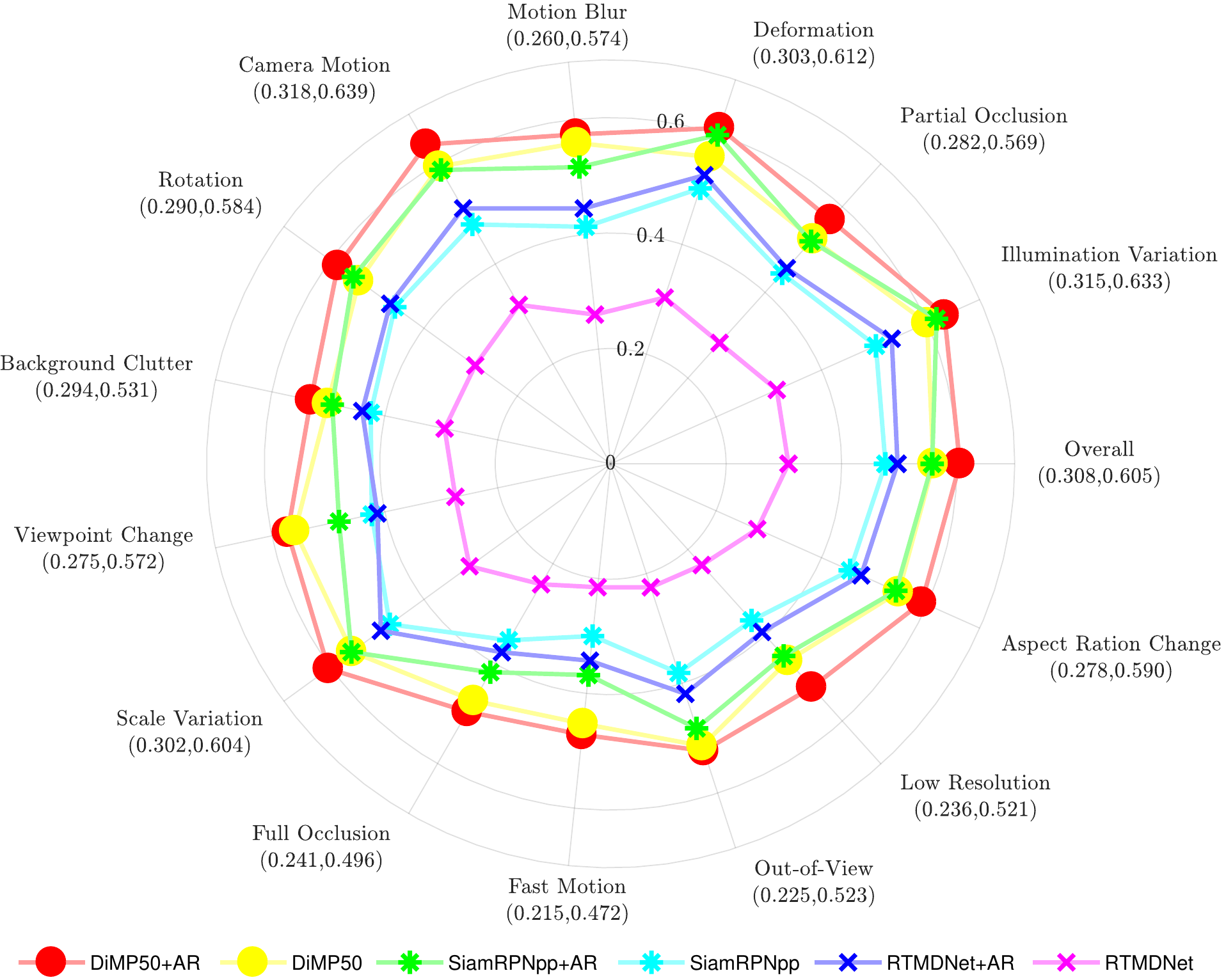}
    \caption{AUC scores of different attributes on the LaSOT \emph{test} set. 
    Alpha-Refine (AR) significantly improve the base trackers 
    on almost all attributes.} \label{fig:lasotatt}
\vspace{-2mm}
\end{figure}
}

\subsection{Ablation Studies}
In this subsection, we conduct ablation studies of our Alpha-Refine (AR) module using 
SiamRPNpp~\cite{SiamRPNplus} as the base tracker, evaluated on the LaSOT~\cite{LaSOT} test 
set. 
% <<<<<<<<<<<<<< Ablation_Head <<<<<<<<<<<<<<<<<<<
\begin{table}[h]
\caption{Analysis of different head options. 
The best three results are marked in \textbf{\textcolor[rgb]{1,0,0}{red}}, \textbf{\textcolor[rgb]{0,1,0}{green}} and \textbf{\textcolor[rgb]{0,0,1}{blue}} bold fonts, respectively.
Numbers are shown in percentage (\%). \label{Ablation_Head}}
\vspace{-2mm}
\begin{center}
\resizebox{0.8\linewidth}{!}{
\begin{tabular}{l|c|c|c}
\hline
\textbf{Method} & $\rm{AUC}$(\%) & $\rm{P_{Norm}}$(\%) &$\rm{P}$(\%) \\
\hline
SiamRPNpp   &47.6  &54.7  &47.2 \\
\hline
+$AR_{rpn} $&50.2  &55.5  &51.2 \\ 
+$AR_{rcnn}$&48.9  &54.2  &46.9 \\ 
+$AR_{c}$   &\textbf{\textcolor[rgb]{0,1,0}{54.6}}   &\textbf{\textcolor[rgb]{0,1,0}{60.3}}   &\textbf{\textcolor[rgb]{0,1,0}{55.3}}  \\
+$AR_{rpn+m}$   &\textbf{\textcolor[rgb]{0,0,1}{53.7}} &\textbf{\textcolor[rgb]{0,0,1}{60.3}} &\textbf{\textcolor[rgb]{0,0,1}{54.7}} \\
+$AR_{rcnn+m}$&51.6&58.1&52.3 \\
+$AR_{c+m}$ &\textbf{\textcolor[rgb]{1,0,0}{55.9}} &\textbf{\textcolor[rgb]{1,0,0}{62.2}} &\textbf{\textcolor[rgb]{1,0,0}{57.4}}  \\
%+$AR_{c+m}^{m2b}$&0.481&0.529&0.503  \\
\hline
\end{tabular}}
\end{center}
\vspace{-4mm}

\end{table}

\noindent{\textbf{Head Options. }}
The head option is a very important component in this work, since it is directly related with the final output. 
Table~\ref{Ablation_Head} reports the performance of the SiamRPNpp+AR tracker with different head options. 
The symbols $\rm{AR}_{rpn}$, $\rm{AR}_{rcnn}$ and $\rm{AR}_{c}$ denote the Alpha-Refine module with RPN style 
box head, RCNN style box head and Key-Point style corner head, respectively.  
`+m' stands for the auxiliary mask head used during training. 
From Table~\ref{Ablation_Head}, we have the following two conclusions: 
(1) all adopted box estimation heads improve the original SiamRPNpp method, and the corner head performs much better than the other two;
and (2) the auxiliary mask head further makes additional improvements, and the combination of the corner and mask heads 
obtains the best performance. 
Thus, we chose $\rm{AR}_{c+m}$ as our final module in this work. 

% <<<<<<<<<<<<<< Ablation_Feature_Fuse <<<<<<<<<<<<<<<<<<<
\begin{table}[htbp]
\caption{Analysis of different feature fusion types. Naive indicates the typical feature 
correlation between reference and test branches. Numbers are shown in percentage (\%).\label{Ablation_FeatFuse}}
\vspace{-2mm}
\begin{center}
\resizebox{0.75\linewidth}{!}{
\begin{tabular}{l|c|c|c}
\hline
\textbf{Method} & $\rm{AUC}$(\%) & $\rm{P_{Norm}}$(\%) &$\rm{P}$(\%) \\
\hline
Pixelwise   &\textbf{\textcolor[rgb]{1,0,0}{55.9}}  &\textbf{\textcolor[rgb]{1,0,0}{62.2}}  &\textbf{\textcolor[rgb]{1,0,0}{57.4}} \\
Depthwise   &54.8  &60.8  &55.8\\
Naive       &53.1  &59.4  &53.9\\ 
\hline
\end{tabular}}
\end{center}
\vspace{-5mm}
\end{table}

\noindent{\textbf{Feature Fusion Options. }}
Table~\ref{Ablation_FeatFuse} compares the {SiamRPNpp+AR} variants using different feature fusion options (naive, depthwise, or pixelwise in Sec~\ref{sec:featfuse}), where the prediction head is determined based on aforementioned discussions.  
The results show that the adopted pixelwise correlation performs the best, indicating that the pixelwise correlation is better at extracting and maintaining spatial information than 
the depthwise correlation or naive correlation.

\begin{table}[h]
\caption{Comparison of different refinement modules. \label{tab:other-refine}
The best result is marked in \textbf{\textcolor[rgb]{1,0,0}{red}} bold fonts.}
\vspace{-2mm}
\begin{center}
\resizebox{0.78\linewidth}{!}{
\begin{tabular}{l|c|c|c}
\hline
\textbf{Method} & $\rm{AUC}$(\%) & $\rm{P_{Norm}}$(\%) &$\rm{P}$(\%) \\
\hline
SiamRPNpp &47.6 &54.7 &47.2 \\
\hline
+IoU-Net  &48.8  &55.6  &47.8\\
+SiamMask &50.3  &54.7  &48.7\\
+AR   &\textbf{\textcolor[rgb]{1,0,0}{55.9}}     &\textbf{\textcolor[rgb]{1,0,0}{62.2}} &\textbf{\textcolor[rgb]{1,0,0}{57.4}}\\
\hline
\end{tabular}}
\end{center}
\vspace{-4mm}
\end{table}

\noindent{\textbf{Comparison with Different Refinement Modules. }} 
We compare our Alpha-Refine (AR) with two recent refinement modules (IoU-Net presented in~\cite{ATOM,DiMP} 
and SiamMask proposed in~\cite{SiamMask}), and report the results in Table~\ref{tab:other-refine}. 
Our Alpha-Refine module surpasses IoU-Net and SiamMask by a large margin.
%
%\emph{More detailed discussions are presented in the supplementary material.} 
%(TODO: introducing AR trained with lite version of data in supplementary)

\noindent{\textbf{Different Backbones. }} 
We investigate the Alpha-Refine module with different backbones and reports the comparison results in 
Table~\ref{Ablation_Backbone}. 
When the ResNet-18 backbone is used, the latency of our AR model is very low but the corresponding 
performance is also 7.4\% higher than the original SiamPRNpp.
As the backbone goes deeper, the AUC score is better but the speed is slower. In this work, we choose 
ResNet-34 as the default backbone to balance accuracy and speed. 
%
%\emph{More speed comparisons are presented in the supplementary material.}
% <<<<<<<<<<<<<< Ablation_Backbone <<<<<<<<<<<<<<<<<<<
\begin{table}[h]
\caption{Accuracy and Speed Comparison of SiamRPNpp+AR with different backbones. \label{Ablation_Backbone}}
\vspace{-2mm}
\begin{center}
\resizebox{0.95\linewidth}{!}{
\begin{tabular}{l|c|c|c|c}
\hline
\textbf{Method}&{\rm AUC}(\%) &fps &latency &\multirow{2}{*}{$\Delta t$} \\
\cline{1-4}
SiamRPNpp         &47.6  &67.1   &14.9ms    &  \\
\hline
+ AR(ResNet-50)   &56.2  &46.5   &21.5ms    &6.6ms  \\
+ AR(ResNet-34)   &55.9  &50.0   &20.0ms    &5.1ms\\
+ AR(ResNet-18)   &55.0  &52.4   &19.1ms    &4.2ms\\
\hline
\end{tabular}}
\end{center}
\vspace{-6mm}
\end{table}

\subsection{Evaluation on Other Benchmarks}
\label{sec_sota}
{\noindent \textbf{TrackingNet.}}
TrackingNet~\cite{Trackingnet} is a popular large-scale short-term tracking benchmark. 
We evaluate various methods on its \emph{test} set, which contains $511$ sequences.
For the \emph{test} set, only groundtruth of the first frame is given and participants 
need to submit their results to the evaluation server.  
Table~\ref{tab:trackingnet} shows that our Alpha-Refine module improves different  
base trackers by a large margin. 
ARDiMPsuper obtains $80.5\%$ in the main $\rm{AUC}$ metric, which is slightly worse 
than the previous best tracker (Siam R-CNN~\cite{SiamRCNN}: 81.2\% in $\rm{AUC}$). 
However, ARDiMPsuper runs approximately $32.9$ \emph{fps}, being six times faster than Siam R-CNN. 

% Table: Comparison on TrackingNet
\begin{table}[!t]
    \caption{Comparison results on the \textit{TrackingNet test} set. 
    `Base': the base tracker; and `Base+AR': the base tracker with Alpha-Refine. 
    The best three results are marked in \textbf{\textcolor[rgb]{1,0,0}{red}}, \textbf{\textcolor[rgb]{0,1,0}{green}} and \textbf{\textcolor[rgb]{0,0,1}{blue}} bold fonts, 
    respectively. Numbers are shown in percentage (\%).}
    \centering
    \resizebox{\linewidth}{!}{
    \begin{tabular}{c|ccc|ccc}
        \hline
        \multirow{2}{*}{\textbf{Method}}&\multicolumn{3}{c|}{\textbf{Base}} &\multicolumn{3}{c}{\textbf{Base+AR}} \\
            \cline{2-7}
            &\rm{AUC}&$\rm{P_{Norm}}$&$\rm{P}$ &$\rm{AUC}$ &$\rm{P_{Norm}}$ &$\rm{P}$ \\
            \hline
        ECO         &61.2  &71.0  &55.9      &75.1  &80.0  &71.4 \\
        RT-MDNet    &58.4  &69.4  &53.3      &76.0  &81.0  &72.3 \\
        ATOM        &70.3  &77.1  &64.8      &77.7  &82.5  &74.5 \\
        SiamRPNpp   &\textbf{\textcolor[rgb]{0,0,1}{73.3}}  &\textbf{\textcolor[rgb]{0,0,1}{80.0}}  &\textbf{\textcolor[rgb]{0,1,0}{69.4}}      &\textbf{\textcolor[rgb]{0,0,1}{78.8}}  &\textbf{\textcolor[rgb]{0,0,1}{83.7}}  &\textbf{\textcolor[rgb]{0,0,1}{76.4}} \\
        DiMP50      &\textbf{\textcolor[rgb]{0,1,0}{74.0}}  &\textbf{\textcolor[rgb]{0,1,0}{80.1}}  &\textbf{\textcolor[rgb]{0,0,1}{68.7}}      &\textbf{\textcolor[rgb]{0,1,0}{79.5}}  &\textbf{\textcolor[rgb]{0,1,0}{84.1}}  &\textbf{\textcolor[rgb]{0,1,0}{76.5}} \\
        DiMPsuper   &\textbf{\textcolor[rgb]{1,0,0}{77.6}}  &\textbf{\textcolor[rgb]{1,0,0}{82.5}}  &\textbf{\textcolor[rgb]{1,0,0}{72.6}}      &\textbf{\textcolor[rgb]{1,0,0}{80.5}}  &\textbf{\textcolor[rgb]{1,0,0}{85.6}}  &\textbf{\textcolor[rgb]{1,0,0}{78.3}} \\
        \hline
    \end{tabular}}
    \label{tab:trackingnet}
\end{table}

{\flushleft \textbf{GOT-10K.}} GOT-10K~\cite{GOT10K} is a recent large-scale dataset, which contains 
10K sequences for training and 180 
for testing. We submit the tracking outputs to the official evaluation server and obtain the 
comparison results (\emph{i.e.}, $\rm{AO}$ and $\rm{SR}_{\rm{T}}$) in Table~\ref{tab:got10k}.
On one hand, our Alpha-Refine module consistently and significantly improves the base trackers in all 
evaluation metrics. 
On the other hand, ARDiMPsuper achieves 70.1\% in the main $\rm{AO}$ metric, which performs much 
better and runs much faster than the previous best tracker (Siam R-CNN~\cite{SiamRCNN}: 64.9\% in 
$\rm{AO}$, 72.8\% in $\rm{SR}_{\rm{T}}$, and 59.7\% in SR$_{\rm{0.75}}$). 

% Table: Comparison on got10k
\begin{table}[!htbp]
    \caption{Comparison results on the \textit{GOT-10K test} set. 
    `Base': the base tracker; and `Base+AR': the base tracker with Alpha-Refine. 
    The best three results are marked in \textbf{\textcolor[rgb]{1,0,0}{red}}, 
    \textbf{\textcolor[rgb]{0,1,0}{green}} and \textbf{\textcolor[rgb]{0,0,1}{blue}} bold fonts, 
    respectively. Numbers are shown in percentage (\%).}
    \centering
    \resizebox{\linewidth}{!}{
    \begin{tabular}{c|ccc|ccc}
        \hline
        \multirow{2}{*}{\textbf{Method}}&\multicolumn{3}{c|}{\textbf{Base}} &\multicolumn{3}{c}{\textbf{Base+AR}} \\
            \cline{2-7}
                    &$\rm{AO}$    &$\rm{SR}_{0.5}$ &$\rm{SR}_{0.75}$  &$\rm{AO}$    &$\rm{SR}_{0.5}$ &$\rm{SR}_{0.75}$ \\
            \hline
        ECO         &41.3  &43.8  &13.4      &56.7  &64.8  &46.1 \\
        RT-MDNet    &35.0  &35.8  &9.2       &56.1  &63.7  &46.9 \\
        ATOM        &\textbf{\textcolor[rgb]{0,0,1}{53.5}}  &\textbf{\textcolor[rgb]{0,0,1}{62.2}}  &\textbf{\textcolor[rgb]{0,0,1}{37.8}}      &\textbf{\textcolor[rgb]{0,0,1}{63.1}}  &\textbf{\textcolor[rgb]{0,0,1}{71.1}}  &\textbf{\textcolor[rgb]{0,0,1}{55.8}} \\
        SiamRPNpp   &51.8  &61.7  &32.4      &61.5  &69.6  &46.9 \\
        DiMP50      &\textbf{\textcolor[rgb]{0,1,0}{60.3}}  &\textbf{\textcolor[rgb]{0,1,0}{71.8}}  &\textbf{\textcolor[rgb]{0,1,0}{46.0}}      &\textbf{\textcolor[rgb]{0,1,0}{65.4}}  &\textbf{\textcolor[rgb]{0,1,0}{74.3}}  &\textbf{\textcolor[rgb]{0,1,0}{58.5}} \\
        DiMPsuper   &\textbf{\textcolor[rgb]{1,0,0}{67.2}}  &\textbf{\textcolor[rgb]{1,0,0}{78.8}}  &\textbf{\textcolor[rgb]{1,0,0}{59.3}}      &\textbf{\textcolor[rgb]{1,0,0}{70.1}}  &\textbf{\textcolor[rgb]{1,0,0}{80.0}}  &\textbf{\textcolor[rgb]{1,0,0}{64.2}} \\
        \hline
    \end{tabular}}
    \label{tab:got10k}
\end{table}

\begin{figure}[htbp]
\centering
    \includegraphics[width=1.0\linewidth]{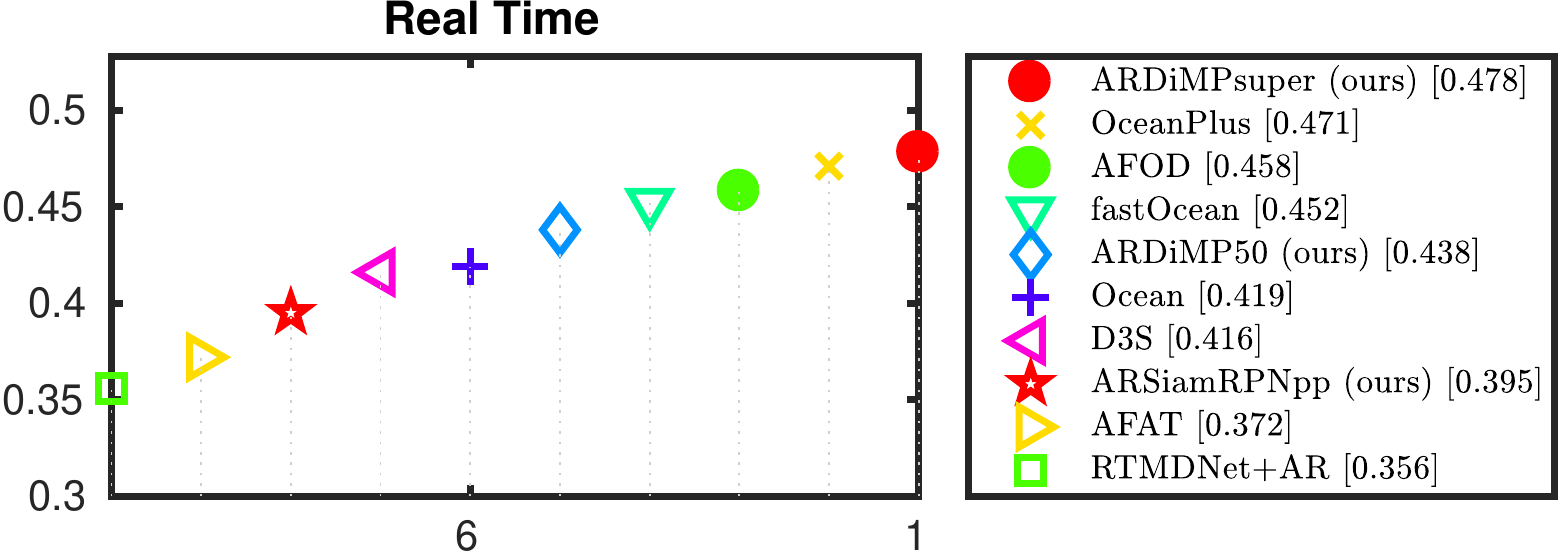}
    \caption{State-of-the-art evaluation on VOT2020. Our ARDiMPsuper obtains the best result in the real-time setting. 
    % >>>
    % , and also achieves top performance in the baseline setting.
    } \label{fig:vot2020-rank-rt}

\end{figure}

{\flushleft \textbf{VOT2020.}} 
VOT2020~\cite{VOT2020} includes $60$ challenging videos with high-quality mask-based ground truth. 
This benchmark takes expected average overlap (EAO) as the main ranking metric, 
which simultaneously considers the trackers' accuracy and robustness. 
The evaluation on VOT2020 has two settings: {\rm{Baseline}} and {\rm{real-time}}. 
The {\rm{real-time}} requires the trackers to predict bounding boxes no slower than the video frame rate (20 fps in the official toolkit). 
The results of different trackers are shown in Table~\ref{tab-vot20}. 
We can see that the proposed Alpha-Refine module improves the base trackers in terms of EAO significantly. 
Besides, Figure~\ref{fig:vot2020-rank-rt} demonstrates that our AR strengthened method ARDiMPsuper obtains the best performance 
in the {\rm{Real-Time}} setting.
\ignore{
and also achieves a promising results (top 4) in the {\rm{Baseline}} setting\footnote{For VOT2020, 
we simply equip the DiMPsuper method with our Alpha-Refine module, without any additional trick.}. 
%\emph{More comparisons and discussions are provided in the supplementary material.}
}

\begin{table}[!t]
    \caption{Comparison results on the VOT2020 benchmark. 
    `Base': the base tracker; and `Base+AR': the base tracker with Alpha-Refine. 
    The best three results are marked in \textbf{\textcolor[rgb]{1,0,0}{red}}, 
    \textbf{\textcolor[rgb]{0,1,0}{green}} and \textbf{\textcolor[rgb]{0,0,1}{blue}} bold fonts, 
    respectively. The main EAO metric is reported.}
    \centering
    \resizebox{\linewidth}{!}{
    \begin{tabular}{c|cc|cc}
        \hline
            \multirow{2}{*}{\textbf{Method}}&\multicolumn{2}{c|}{\textbf{Base}} &\multicolumn{2}{c}{\textbf{Base+AR}} \\
            \cline{2-5}
            &Baseline   &Real Time &Baseline   &Real Time \\
            \hline
        RT-MDNet    &0.248  &0.247      &0.371  &0.356 \\
        SiamRPNpp   &0.254  &0.254      &0.395  &0.395 \\
        ECO         &\textbf{\textcolor[rgb]{0,0,1}{0.280}}  &\textbf{\textcolor[rgb]{0,0,1}{0.276}}      &\textbf{\textcolor[rgb]{0,0,1}{0.426}}  &\textbf{\textcolor[rgb]{0,0,1}{0.426}} \\
        ATOM        &0.275  &0.279      &0.416  &0.414 \\
        DiMP50      &\textbf{\textcolor[rgb]{0,1,0}{0.286}}  &\textbf{\textcolor[rgb]{0,1,0}{0.278}}        &\textbf{\textcolor[rgb]{0,1,0}{0.444}}  &\textbf{\textcolor[rgb]{0,1,0}{0.438}} \\
        DiMPsuper   &\textbf{\textcolor[rgb]{1,0,0}{0.314}}  &\textbf{\textcolor[rgb]{1,0,0}{0.311}}      &\textbf{\textcolor[rgb]{1,0,0}{0.471}}  &\textbf{\textcolor[rgb]{1,0,0}{0.478}} \\
        \hline
    \end{tabular}}
    % \begin{tabular}{ccccccc}
    %     \hline
    %         &DiMPSuper&DiMP50&ATOM&ECO&SiamRPN&RTMDNet \\
    %         \hline
    % Baseline&0.314  &0.286  &0.275  &0.280  &0.254  &0.248 \\
    % RealTime&0.311  &0.278  &0.279  &0.276  &0.254  &0.247 \\
    %         \hline
    %         &DiMPSuperAR34a&DiMP50AR34a&ATOMAR34a&ECOAR34a&SiamRPNAR34a&RTMDNetR34a \\
    %         \hline
    % Baseline&0.471  &0.444  &0.416  &0.426  &0.395  &0.371 \\
    % RealTime&0.478  &0.438  &0.414  &0.426  &0.395  &0.356 \\
    %     \hline
    % \end{tabular}
    \label{tab-vot20}

\end{table}

%\subsection{Discussion} (it in supplementary)
%\paragraph{Can Alpha-Refine track independently?} We evaluate Alpha-Refine as a tracker on LaSOT Dataset. Alpha-Refine achieve a performance of 0.456 AUC score by itself, which is not. (TODO: Comment and analyze this issue) 
%\textcolor{red}{Move this section to supplement!}
%\textcolor{red}{When using VOT2020, I guess the Alpha-Refine achieves high accuracy, the base tracker achieves high robustness, they are complementary!! So I think it is very interesting on this point.}

\section{Conclusion.}
In this work, we propose a novel Alpha-Refine method for visual tracking, which is an accurate and general 
refinement module to effectively improve the tracking performance of different types of trackers in a plug-and-play style. 
By exploring multiple design options, we find that extracting and maintaining precise spatial information is the key to the precise box estimation. Alpha-Refine finally adopts a precise pixel-wise correlation layer, a Key-Point style prediction head, 
and an auxiliary mask head. Finally, we apply the Alpha-Refine model to six well-known and top-performed trackers and conduct 
numerous evaluations on four popular benchmarks. The experimental results demonstrate that our Alpha-Refine could consistently 
improve the tracking performance with few computational loads.
\vspace{4mm}
%------------------------------------------------------------------------

\noindent{\textbf{Acknowledgement. }}This work was supported in part by the National Natural Science Foundation of China 
under Grant nos. 62022021, 61806037, 61872056, and 61725202, and in part by the Science and Technology Innovation 
Foundation of Dalian under Grant no. 2020JJ26GX036.

\clearpage
{\small
\bibliographystyle{ieee_fullname}
\bibliography{egbib}
}

\end{document}